%% file: main.tex
\documentclass{article}
\usepackage[accepted]{aistats2020}

\usepackage{times}
\usepackage{soul}
\usepackage{url}
\usepackage[hidelinks]{hyperref}
\usepackage[utf8]{inputenc}
\usepackage[small]{caption}
\usepackage{graphicx}
\usepackage{amsmath}
\usepackage{amsthm}
\usepackage{booktabs}
\usepackage{xcolor}
\usepackage[inline]{enumitem}
\usepackage{appendix}

\usepackage{booktabs}       
\usepackage{amsfonts}       

\usepackage{graphicx}
\usepackage{hyperref}
\usepackage{subcaption, caption}

\usepackage{appendix}

\usepackage{booktabs}       
\usepackage{amsfonts}       

\usepackage{graphicx}
\usepackage{hyperref}

\graphicspath{ {./imgs/} }
\usepackage{mathtools}
\usepackage{wrapfig}
\usepackage{subcaption, caption} 
\usepackage[round]{natbib}
\usepackage{nicefrac}
\usepackage{float}

\usepackage{algorithm} 
\usepackage{algpseudocode}
\floatname{algorithm}{Procedure}

\title{Graph Convolutional Policy for Solving Tree Decomposition via Reinforcement Learning Heuristics}

\author{
    Author
    \affiliations
    Institution
}

\begin{document}


\twocolumn[

\aistatstitle{Graph Convolutional Policy for Solving Tree Decomposition via Reinforcement Learning Heuristics}

\aistatsauthor{  Taras Khakhulin \And
    Roman Schutski 
    \And Ivan Oseledets }

\aistatsaddress{ taras.khakhulin@skoltech.ru \And r.schutski@skoltech.ru \And i.oseledets@skoltech.ru}
\vspace{-10pt}
\aistatsinstitute{ Skolkovo Institute of Science and Technology, Moscow} ]

\begin{abstract}
We propose a Reinforcement Learning-based approach to approximately solve the Tree Decomposition (TD) problem. TD is a combinatorial problem, which is central to the analysis of graph minor structure and computational complexity, as well as in the algorithms of probabilistic inference, register allocation, and other practical tasks. 
Recently, it has been shown that combinatorial problems can be successfully solved by learned heuristics. 
Our model is based on the graph convolution neural network (GCN) for learning graph representations and the Actor-Critic method for training. We establish that this simple approach successfully generalizes from small graphs, where TD can be found by exact algorithms, to large instances of practical interest, while still having very low time-to-solution. On the other hand, the agent-based approach surpasses all classical greedy heuristics by the quality of the solution. Surprisingly, we also find that even a single graph is sufficient to train the heuristic, which may suggest a high regularity of the structure of the Tree Decomposition problem.
\end{abstract}

\section{Introduction}
At the core of many practical tasks, such as probabilistic inference, decision making, planning, etc., lies a combinatorial optimization problem. The solution of large NP-problems is often possible only with the help of heuristics.
These heuristics are designed manually, which is a complicated and time-consuming process. The resulting algorithm is also typically domain-specific and can not be reused. Recently, an application of Reinforcement Learning (RL) to design heuristics gained significant attention ~\cite{BelloPLNB16,Weilling_routing18,DaiKZDS17}. RL is a natural framework for the automatic design of approximation algorithms for problems with an inherent cost function and large search space, which is the essence of combinatorial optimization.

The specific NP problem we consider here is the \emph{Tree Decomposition}, first introduced \citet{robertson1986graph}. The Tree Decomposition (TD) is central to the analysis of the complexity and the topological structure of graphs. 
Also, if the TD of a graph is known, then several NP-hard problems can be solved in linear time using it. Examples are Independent Set, Clique, Satisfiability, Graph Coloring, Travelling Salesman Problem (TSP), and many others.
The solution of the TD problem is characterized by an integer parameter treewidth. The treewidth quantifies the complexity of many NP-problems; the computational cost of solving these problems is exponential in the treewidth, but only polynomial in the problem's graph size.
Tree Decomposition emerges as a core step in various contexts, such as probabilistic inference \citep{Kask2011PushingTP} or shortest path search \citep{shortest_path_tw}. The TD problem is usually solved on non-Euclidean graphs, as opposed to the traveling salesman problem (TSP), which is the most common target of recent trainable heuristics studies \cite{Weilling_routing18,BelloPLNB16}.

Several exact \cite{Gogate_quickbb12,tamaki2019positive,bodlaender2006exact} and approximate \cite{BerryMinDegree} algorithms exist to solve the TD problem. We are the first (to the best of our knowledge) to propose a Reinforcement Learning based solution. 
To learn heuristics, we utilize Markov Decision Process (MDP) formalism, considering TD as a Reinforcement Learning task with graph-structured data.
We demonstrate that the resulting policy can successfully \emph{generalize} to problems with different graph structures and sizes. Our findings show that the agent can be trained even on a \emph{single graph}.
The quality of the solution of our agent-based procedure is superior compared to all simple greedy heuristics, and the time-to-solution is much lower compared to advanced algorithms.
 
 The goal of this work is not to outperform the existing state of the art TD algorithms, but to provide a direction in the study of RL approaches to fundamental NP-hard problems.

Main contributions of the paper are:
 \begin{itemize}
    \item A method is proposed to learn the heuristic for Tree Decomposition Problem, which is more accurate than simple polynomial solvers and has similar low time-to-solution.
    \item It is shown that the agent trained on a small single graph generalizes to large real-world instances of TD problem and preserves high quality of the solution.
    \item We demonstrate that our stochastic policy generalizes better across different graph structures.
 \end{itemize}

\section{Background}
\input{parts/background.tex}
\section{Method}
\input{parts/method.tex}
\input{parts/experiments.tex}
\input{parts/related_works.tex}
\input{parts/conclusion.tex}

\newpage
\newpage

\bibliographystyle{named}
\bibliography{literature.bib}
\newpage
\input{parts/appendix}
\end{document}

%% file: parts/background.tex
In this section, we introduce the problem and the techniques we use. We start with a formulation of Tree Decomposition as a linear ordering problem. Then we explain the embedding method using Graph Convolutional Networks~\citep{KipfW16}. Finally, we formulate the problem in the Reinforcement Learning framework.
\subsection{Tree decomposition problem}
A full definition of Tree Decomposition can be found in the original work by \citep{robertson1986graph} or in more recent review~\citep{bodlaender1994tourist}. 
Informally, the Tree Decomposition measures how close a given graph resembles a tree.

Formally, a Tree Decomposition is a mapping
 of the initial graph $G = (U, E)$ into a tree graph $F = (B, T)$. Here $U$ is the set of nodes, and $E$ is the set of edges of the initial graph; $B$ is the set of \emph{bags} (each bag $b \in B$ is a subset of nodes of the graph $G$, $b \subset U$) and $T$ is the set of the edges of the tree graph. A Tree Decomposition has to fulfill three criteria to be valid:
\begin{enumerate}
\item Every node of $G$ is in some bag, i.e., $\cup_{b \in B} b = U$.
\item For every edge $(u, v) \in E$ there must be a bag such that both endpoints are in that bag, i.e., $\exists b: u \in b, v \in b$.
\item For every node $u$ of $G$, the subgraph of the tree $F$, induced by all bags that contain $u$ is a connected tree.
\end{enumerate}
The Tree Decomposition problem is as follows: For a given graph $G$, one needs to find a tree $F$ satisfying the conditions above, such that the size of the maximal bag $b \in B$ minus one, called \emph{treewidth}, is minimized across all possible tree graphs. It can be shown that this problem is NP-hard \citep{finnish_article}. Instead of building the tree $F$ directly, in this work we will search for TD by using its relation to the ordering of vertices \citep{clique_trees}.  The procedure to build a tree $F$ given a permutation of vertices is described in Appendix \ref{app:td_from_order}. 

A permutation \(\pi(u): u \in U \to [1 \dots |U|]\) of vertices of a graph $G = (U, E)$ is called an \textit{elimination} order. Consequently, $\pi^{-1}(t): t \in [1 \dots |U|] \to U$ is an inverse function of the order. Given a number $\pi^{-1}(t)$ returns a node. The elimination order of the graph yields the following procedure:

\begin{enumerate}
    \item For $t \in [1\dots|U|]$, take the $t$-th node $u = \pi^{-1}(t)$.
    \item Remove $u$ and connect all neighbors of $u$ into a clique (fully connected subgraph). 
\end{enumerate}

If $G$ has treewidth at most \(k\), then there is an elimination order \(\pi\) of \(G,\) such that each vertex has at most \(k\) neighbors in the elimination procedure with respect to \(\pi\) \citep{Amir2001EfficientAF}.
We define the maximal number of neighbors associated with a permutation $\pi$ as $c_{\pi}$. The treewidth is a minimum of $c_{\pi}$ across all possible permutations, i.e. $tw(G) = \underset{\pi}{\min} ~ c_\pi$.
If the treewidth of a graph is small, then it is tree-like. In particular, a tree has treewidth 1.

We define our problem as follows: given an undirected graph $G$, find an elimination order, i.e., a permutation of the vertices, such that the number of neighbors in the elimination procedure along $\pi$ is minimal across all permutations.

\begin{figure}[t]
    \centering
    \includegraphics[width=0.45\textwidth]{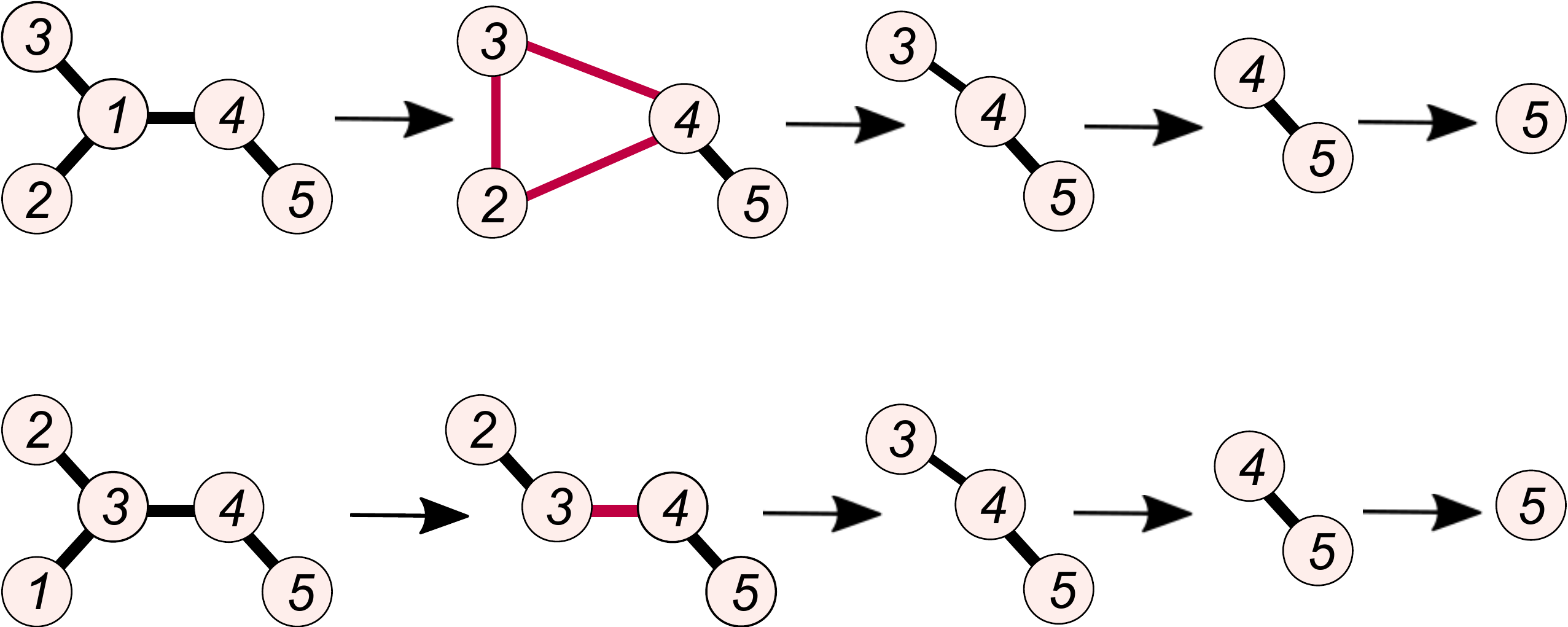}
    \caption{Elimination procedure for a given order of nodes $\pi$. The labels on nodes correspond to their order. In red is shown a maximal clique that emerges during the elimination procedure. The second order is optimal: the clique size is minimized}
    \label{fig:fillin}
\end{figure}

As an example, consider two elimination procedures of the same graph in Figure \ref{fig:fillin}. In the first case (upper part), the size of the maximal clique is 3 (and hence the treewidth corresponding to this order is 2). In the second case, the treewidth is 1, and the order is optimal (since the graph is a tree). Note that the optimal order may not be unique: the order of nodes number 1 and 2 can be swapped, for example, but the same treewidth is achieved.

\subsection{Graph Convolutional Networks} 
Our problem has a non-Euclidean domain, and we need a neural network that is efficient for graphs with arbitrary structures. One of the most popular options are graph convolutional networks.

Recent graph architectures follow the aggregation scheme, which consists of three types of functions: message passing, aggregation, and update function. In GCN   \citep{KipfW16,DefferrardBV16}, message passing is expressed as a simple multiplication of features by weights; aggregation is done by summation, and update is an activation function. Let $\mathbf{A}$ be an adjacency matrix of a graph of $n$ vertices, let $\tilde{\mathbf{A}} = \mathbf{A} + \mathbf{I}_n$  be an adjacency matrix with self-connections, and let $\mathbf{H}^{(0)}$ be a feature matrix defined on the nodes of $G$. The propagation rule of a GCN layer is:
\begin{equation}
    f\left(\mathbf{H}^{(l)}, \mathbf{A}\right)=\sigma\left(\mathbf{D}^{-\frac{1}{2}} \tilde{\mathbf{A}} \mathbf{D}^{-\frac{1}{2}} \mathbf{H}^{(l)} \mathbf{W}^{(l)}\right)
\end{equation}
Here \( \mathbf{D}\) is a diagonal matrix with elements defined as \(\mathbf{D}_{i i}=\sum_{j} \tilde{\mathbf{A}}_{i j} \),
$\mathbf{H}^{(l)}$ is a feature matrix at $l$-th layer, $\mathbf{W}^{(l)}$ are the weights of the $l$-th layer and $\sigma$ is an activation function.
After $l$ layers, the features of GCN contain information about $l$-hop neighbors in $G$.


\subsection{Reinforcement Learning for Graph Elimination}

To formulate the TD problem in the RL framework, we should define a Markov decision process (MDP) for the agent. At each time step $t$, the agent selects a node of the graph $G_{t}$ based on the observed information, represented by the state features. The node is then eliminated from $G_{t}$ and the graph $G_{t+1}$ is produced. We define the MDP $( \mathcal{S}, \mathcal{A}, P, r, \gamma)$ as follows:
\begin{itemize}[label=\textbullet, nolistsep]
    \item $\mathcal{S}$ is a set of states. Each state $s_{t} \in \mathcal{S}$ is defined as the embedding matrix $\mathbf{H}^{(l)}$ of the graph $G_{t}$. We use GCN, to extract features on every step. 
    \item $ \mathcal{A}$ is a set of actions. It consists of nodes $u \in U_{t} \subset U$, which have not been eliminated at the current step.
    \item  $P$ is a transition function \(P=p\left(s, u, s^{\prime}\right),\) where \(p\left(s, u, s^{\prime}\right)\) is a probability distribution, \(s, s^{\prime} \in \mathcal{S}, u \in \mathcal{A} .\)
    \item $r$ is the cost for the sequential combinatorial problem; in our case, $r$ is the size of a maximal clique in the graph, which is obtained during the elimination process. This reward provides an estimation of the treewidth for the current elimination order.
    \item $\gamma \in [0, 1]$ is a discount factor responsible for trading off the preferences between a current reward and a future reward.
\end{itemize}


%% file: parts/method.tex




\subsection{Graph Convolutional Policy}

We represent graph $G_{t}$ with $n$ nodes as the pair $(\mathbf{A_{t}}, \mathbf{H_{t}})$, where $\mathbf{A_{t}}$ is the adjacency matrix and $\mathbf{H_{t}} \in \mathbb{R}^{n \times d}$ is the node feature matrix, assuming that the graph has $d$ features. We define $\mathbf{H_{0}}$ as a vector of the inverse node's degree.



\begin{figure}[h!]
    \centering
    \includegraphics[width=0.48\textwidth]{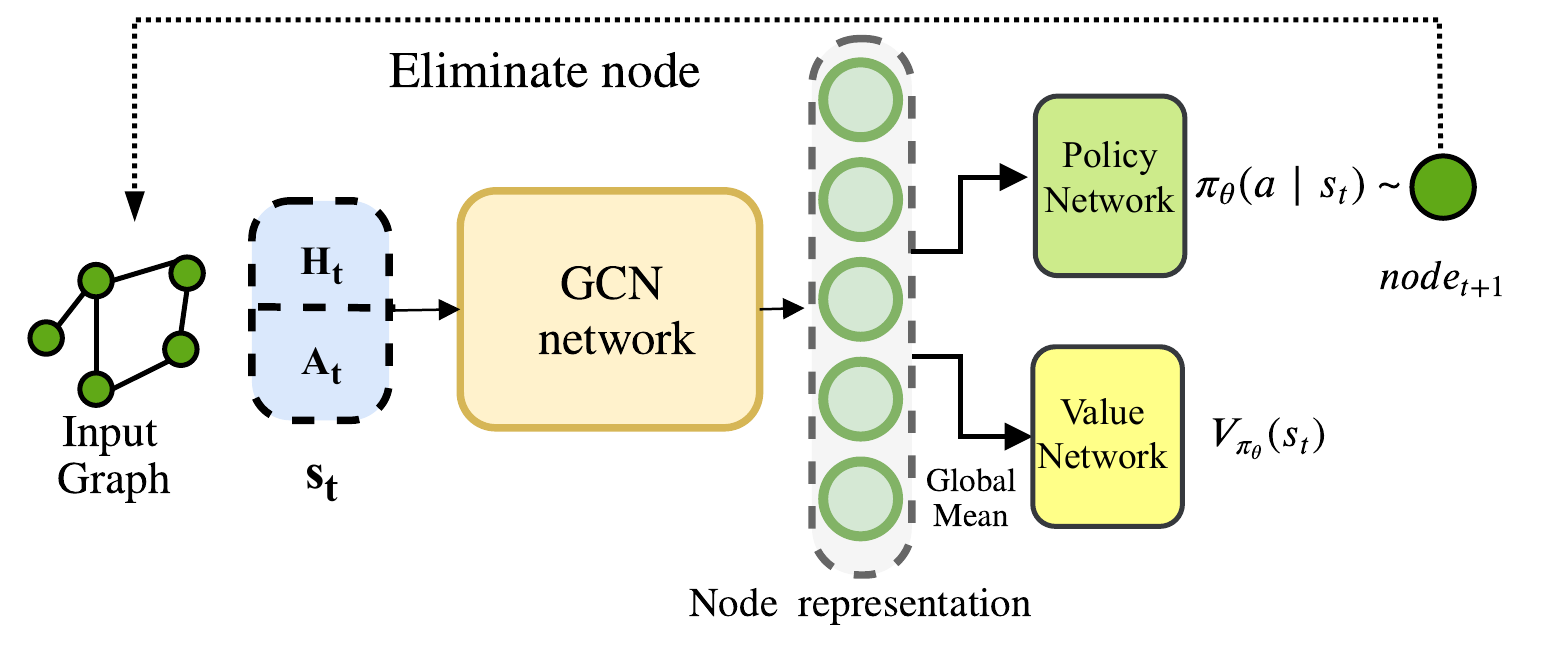}
    \caption{The diagram of the agent-based heuristic. We eliminate node sampled from the policy at every step.}
    \label{fig:gcn_fm}
\end{figure}


In order to work with dense reward signal in the setting of the TD problem, we define a reward function as follows: 
\begin{equation}
    r(s_t,u) = \begin{cases}  -c_t(u), \text{if } t \text{ is terminal,} \\ - \log{c_t(u)} \text{, otherwise,} \end{cases}
    \label{eq:cost_function}
\end{equation} 
where $c_t(u) = \max\{\text{degree}(u), c_{t-1}(u)\}$, and $u$ is the node of the graph $G_{t}$ which defines a state $s_{t}$. Such reward will yield $tw(G)$ in the end of the elimination process (which we are trying to minimize), and at the same time it provides some feedback to the agent during the search of the optimal ordering.

Our node selection policy $\Pi_{\theta}(u|s_t)$ is parameterized by three layers of GCN \citep{KipfW16,DefferrardBV16} with ELU activations \citep{Clevert2015FastAA} and two heads based on two-layer linear networks with ELU for value and policy functions. Here $\theta$ denotes all weights of the network. An attractive property of GCN in our setting is permutation invariance of state representation, which allows us to learn the pattern of heuristic irrespective of the order of the initial adjacency matrix.
Figure \ref{fig:gcn_fm} summarizes our method.

\subsection{Policy Optimization}
We use the Actor-Critic method \citep{Konda:2002} to search for the optimal policy in the MDP, which is explicitly defined above.
Let us briefly describe the Actor-Critic method below.

To find an optimal policy $\Pi_{\theta}$, one needs to maximize the expectation of the reward, defined as: 
\begin{equation}
    L_{\Pi_{\theta}}(s)=\mathbb{E}_{u \sim\Pi_{\theta}(u | s)}[r(s, u)]
\end{equation}
To reduce the variance we use a trainable value function $V_{\Pi_{\theta}}(s)$ as a baseline, which is usual for Actor-Critic methods \citep{Konda:2002}. The purpose of a baseline is to estimate the complexity of the current state $s$. The value function is parameterized as a two-layer perceptron network, the same as the policy network. The value network takes as input a global average of node embeddings and is trained with a standard $L_2$ loss function.  We take the loss of the value network as:
\begin{equation}
\begin{aligned}
&L_{\text{value}}(s,u) =\sum_{t}\left(\hat{A}(s_{t}, u_{t})-V_{\Pi_{\theta}}(s_{t})\right)^2\\
\end{aligned}
\end{equation}
Here \(\hat{A}\left(s_{t}, u_{t}\right)\) is the advantage estimation function. We defined the advantage according to the \emph{Generalized Advantage Estimator} (GAE) method by \citet{Schulman2015HighDimensionalCC}. For a given trajectory $\pi$ for each step $t$, the GAE method implements an exponential average of advantage estimations along with all future steps in the trajectory. The GAE achieves the trade-off between the bias and variance by adjusting the weight $\lambda$ in the exponential average. The estimate of advantage is expressed as:
\begin{equation}
\begin{aligned} 
& \hat{A}^{G A E}\left(s_{t}, u_{t}\right) = \sum_{l=0}^{|U|- t} (\gamma\lambda)^{l} \delta_{t+l},\\
&\delta_{t} =r_{t}+\gamma V\left(s_{t+1}\right)-V\left(s_{t}\right)
\end{aligned}
\end{equation}
We also included an entropy term in the loss function to increase the exploration of our agent. The entropy loss is defined as:
\begin{equation}
    L_{\text{entropy}}(s, u)=-\sum_{t} \Pi_{\theta}(u_t | s_t) \log \Pi_{\theta}(u_t | s_t)
\end{equation}
The total loss function for simultaneous training of graph representation,  policy, and value function is a weighted sum of the terms described above.
\begin{equation}
    L_{\text{total}}(s,u) = \text{L}_{\Pi_{\theta}} + \beta_{\text{value}} L_{\text{value}} - \beta_{\text{entropy}} L_{\text{entropy}}
\end{equation}

%% file: parts/experiments.tex
\section{Experiments}
In this section, we demonstrate that the agent can successively learn a heuristic.  
We present our comparative results against common human-designed heuristics and show that our neural heuristic trained on a single graph can generalize on the graphs with a large number of the vertices. 


    

    


\subsection{Data}
We use three datasets with different structures for the experiments.
The first dataset consists of random Erd\H{o}s–R\'enyi (ER) graphs \citep{Erdos:1960} with edge probability $5/n$, where $n$ is the number of nodes. We experimentally found that this choice of the edge probability leads to the not trivial TD problem (the ER graphs contain many cycles and are not too dense).
For the validation, we use a fixed set $\mathcal{D}$ of 100 Erd\H{o}s–R\'enyi graphs with 10 to 1000 nodes and the set $\mathcal{D}_{\text{small}}$ as first 50 graphs from $\mathcal{D}$. 

The second dataset is taken from the PACE2017 competition \citep{Dell2017TheP2} on designing TD algorithms\footnote{\url{https://github.com/PACE-challenge/Treewidth-PACE-2017-instances}}.

The third dataset comprises graphs that emerge during the simulation of random quantum circuits \citep{boixo2017simulation}, a common framework in the study of quantum computing supremacy.

The main reason for using the ER graphs is to simplify the reproducibility of the experiments. Our main results are obtained on the PACE2017 dataset, which was created specifically to test TD algorithms and contains complicated problem instances. The third dataset is selected as a practical TD application example.


\subsection{Training Details} 
Our code is based on the PyTorch Geometric \citep{Fey/Lenssen/2019}. We train our models with the Adam optimizer \citep{Kingma2014AdamAM} with parameters $\beta_1 = 0.9$, $\beta_2 = 0.999$ and learning rate $lr=0.008$. 
We also set the parameters of the RL algorithm as follows: the discount factor $\gamma = 0.999$, the GAE weight $\lambda = 0.85$, the weight of the value loss $\beta_{\text{value}}=1.0$, and the multiplier of the entropy regularization  $\beta_{\text{entropy}}=0.001$. 
The GCN subnetwork contains 3 layers, and the hidden feature size is 64, as in \citet{DaiKZDS17}. Policy and value heads are parameterized by a 2-layer perceptron with a hidden dimension equal to 64. All hyperparameters are selected with a grid search on the hold-out validation set.
We train our model on the NVIDIA 1080ti with one thread for sampling. 

 \begin{figure}[t]
    \centering
    \vspace{-10pt}
    \includegraphics[width=0.44\textwidth]{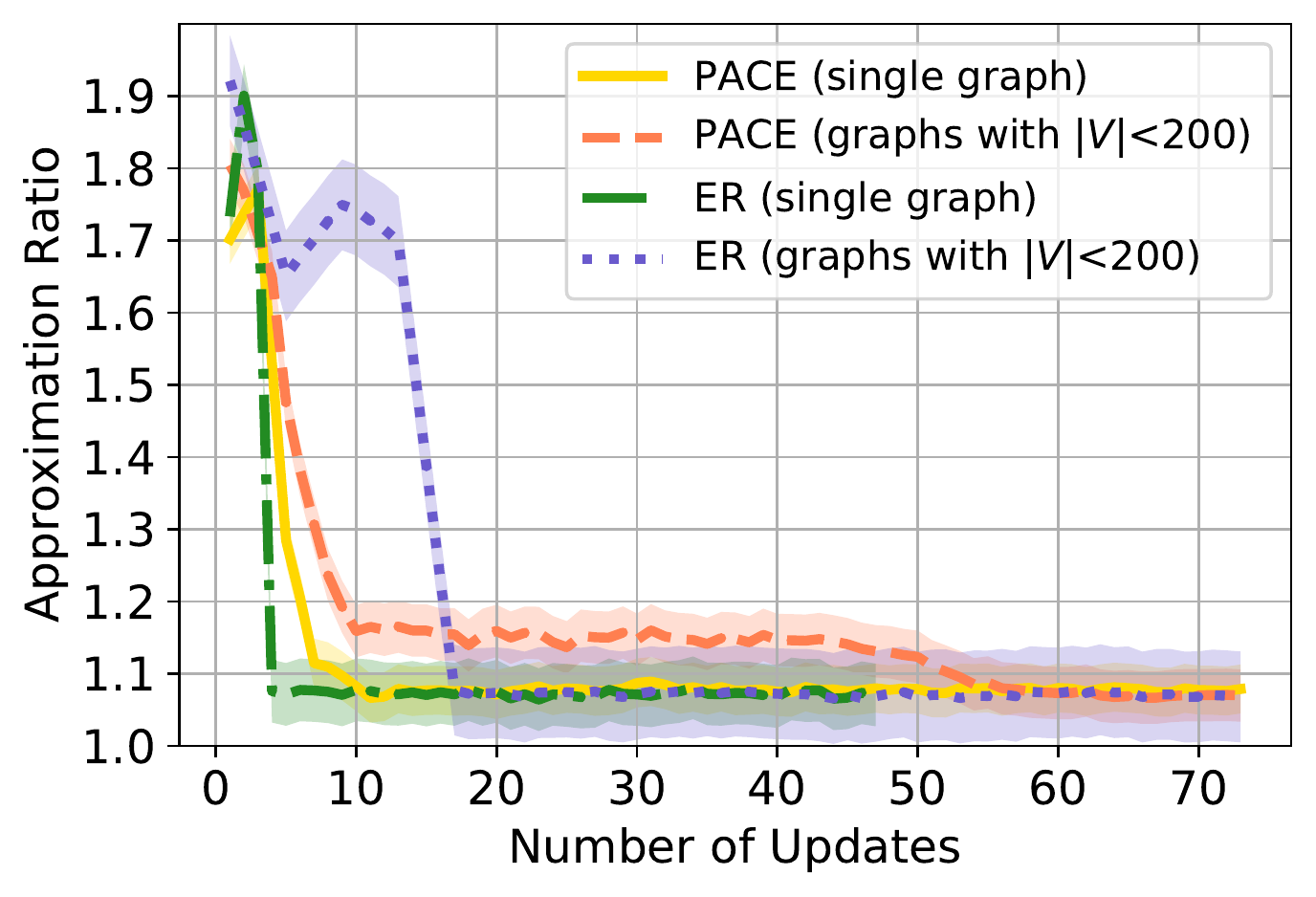}
    \caption{Learning curves showing the Approximation Ratio (less is better) and number of updates (mean and variance over 5 random seeds). The agents are trained on the different sets and compared on the fixed validation set $\mathcal{D}_{\text{small}}$.
    For training on a single graph, we choose the graph "he064" from the PACE dataset with $|V| = 68$, and the ER graph with the same number of nodes. For training on the graphs with $|V|<200$, we sample a new graph at every update step. The agent trained on a single graph achieves the same accuracy in a lower number of training iterations.}
    \label{fig:selection_agent}
    \vspace{-13.8pt}
\end{figure}
\subsection{Baseline and Evaluation Details}
Before starting to experiment with our model, we should introduce other algorithms used in this work. We use two greedy heuristics, two specialized TD solvers, and adapt the S2V-DQN method to the TD problem.

The greedy heuristics are minimal degree and minimal fill-in \citep{bodlaender1994tourist}. They work fast enough and are accurate in practice \citep{bodlaender_comparison}.
The minimal degree heuristic selects nodes with a minimal number of neighbors.
The minimal fill-in algorithm selects the nodes such that the number of introduced edges at each step is minimized.

Two specialized solvers are based on very different approaches, and both will produce an exact solution if provided enough running time (exponential in the graph size). The first solver by \citep{tamaki2019positive} employs the connection of Tree Decomposition and vertex separators. This method searches for optimal TD directly in the space of tree graphs without referring to the ordering of vertices.  It is a winner algorithm on the PACE2017 competition.
Another powerful solver is QuickBB by \citep{Gogate_quickbb12}, which is based on the branch and bound algorithm.  We restrict the runtime of both solvers to 30 minutes per graph instance as in the PACE2017 competition \citep{Dell2017TheP2}.

We also compare with another Reinforcement Learning approach by \citep{DaiKZDS17} called S2V-DQN. This algorithm was previously used to solve sequential optimization problems on graphs, such as Minimum Vertex Cover, and we adapt it to use the cost function from Eq.~\ref{eq:cost_function}. We train the S2V-DQN model\footnote{our implementation with the author's parameters} on a single graph to have faster convergence. 

To produce a solution from our RL-based heuristic, we sample 10 trajectories and take the one with the \emph{lowest} treewidth.


We compare the performance of different solvers with respect to the solver of \citet{tamaki2019positive} work. As a performance metric an \emph{Approximation Ratio} (AR), $\text{AR} = \frac{tw_{\text{method}}(G)}{tw_{\text{tamaki}}(G)}$ is used. The AR metric is the standard in the literature on approximation algorithms. The treewidth is calculated from the elimination order $\pi$ produced by the solvers.

\begin{table}[t]
\centering
\begin{tabular}{l|cccc}
\toprule
 \begin{tabular}[c]{@{}l@{}}  Learnable \\ Heuristic \end{tabular} & 10 & 50 & 100 & Std. Dev. \\
\hline
GCN-agent & 1.08 & 1.09 & 1.09 & 0.024 \\
S2V-DQN & 1.67  & 1.10  & 1.11 & 0.028   \\
\bottomrule
\end{tabular}
\caption{Performance of RL-based algorithms on the dataset $\mathcal{D}$. The agents are trained for 50 and 100 epochs (shown in braces) on 6 single different graphs with 50 nodes. The AR is calculated for all graphs in the hold-out dataset, and averaged over the trained agents.}\label{tab:s2v_over_our}
\end{table}

%


 \begin{figure}[t]
    \centering
    \includegraphics[width=0.4\textwidth]{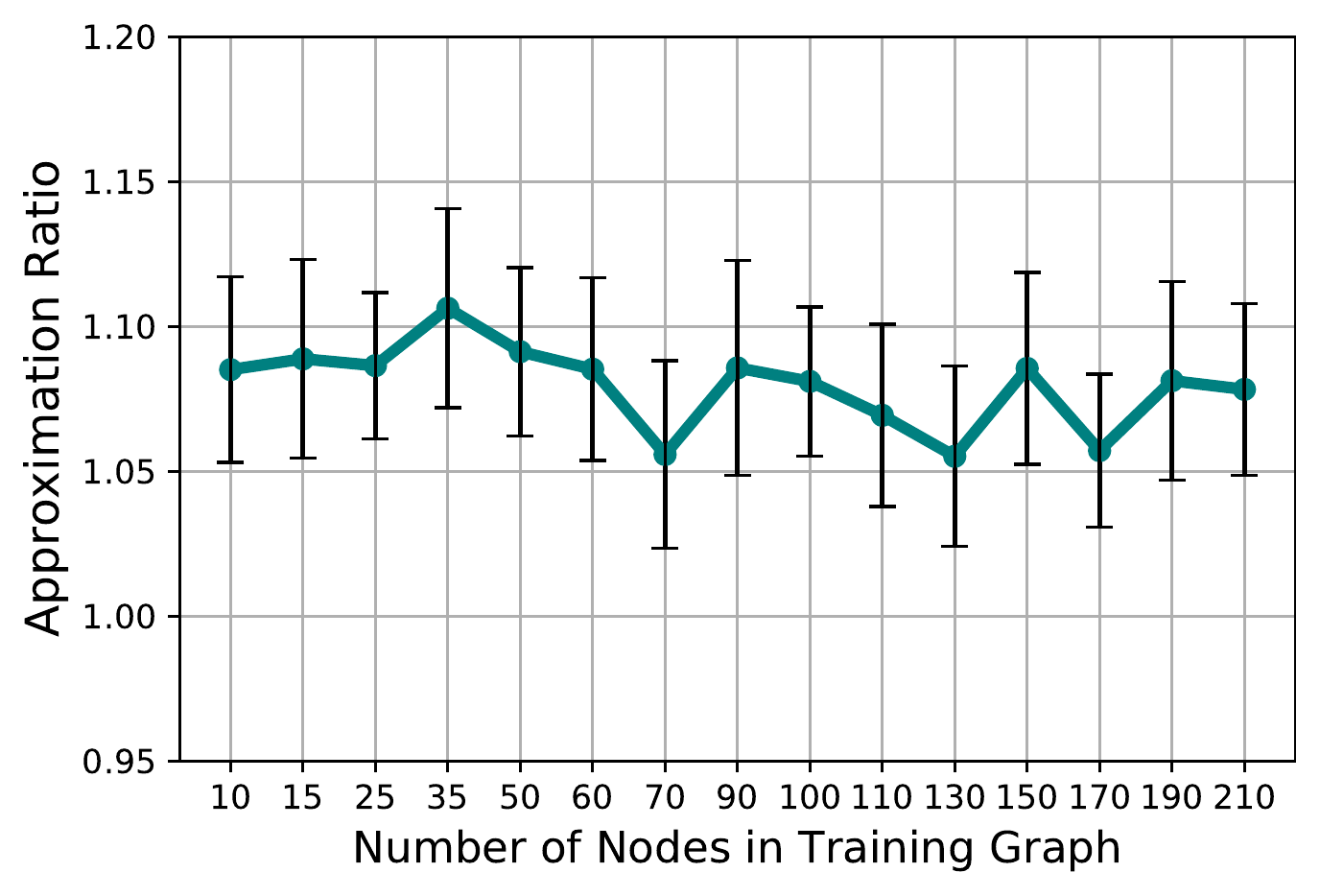}
    \caption{Dependence of the Approximation Ratio (evaluated on the dataset $\mathcal{D}$) on the number of nodes in the training graph.
    The validation results slightly vary on the size of a training graph. }
    \label{fig:selection_graph_size}
\end{figure}
 
\begin{figure*}[h]
     \centering
     \begin{subfigure}[b]{0.32\textwidth}
         \centering
         \includegraphics[width=\textwidth]{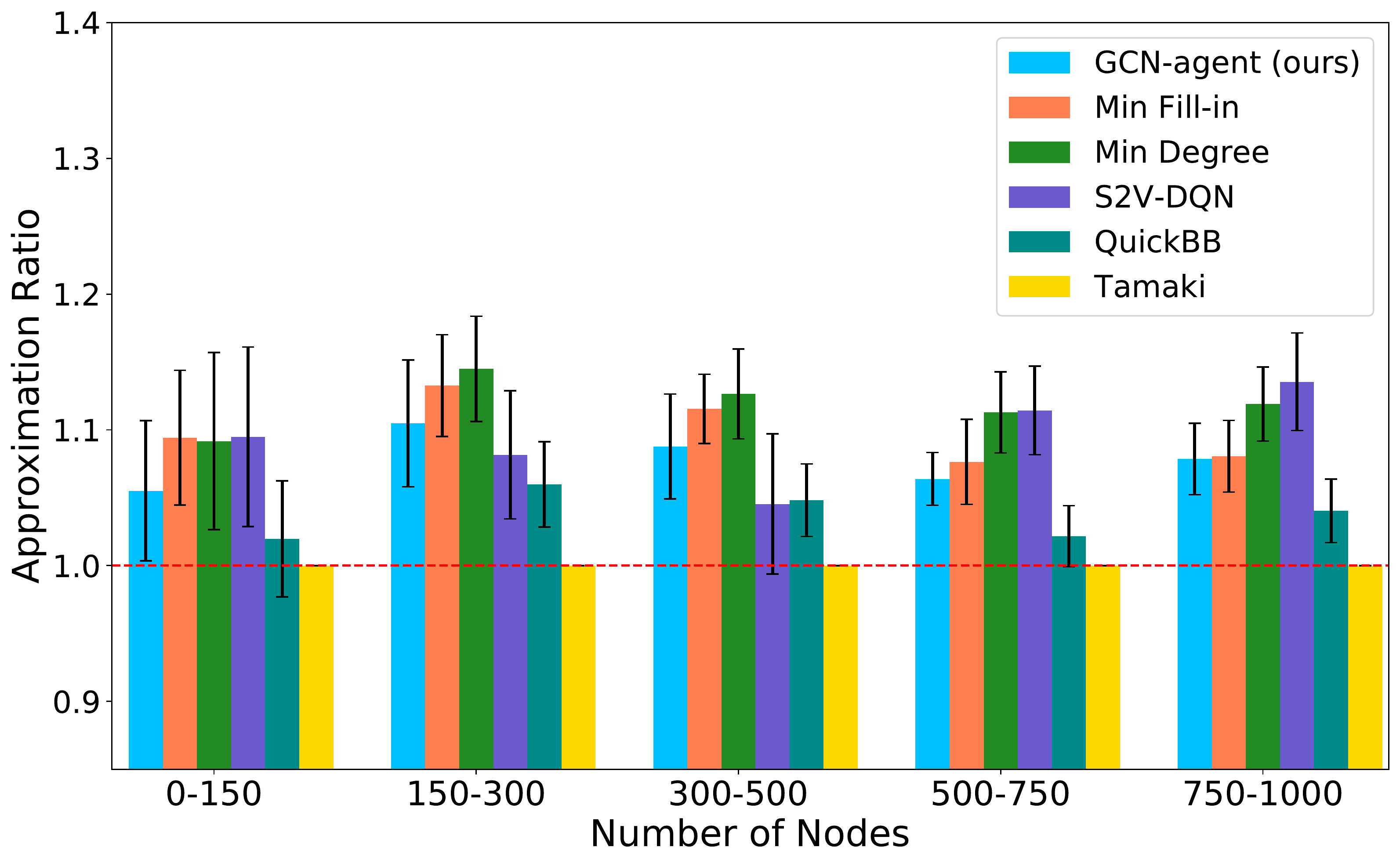}
         \caption{Erd\H{o}s–R\'enyi}
         \label{fig:erdos_bar}
     \end{subfigure}
     \hfill
     \begin{subfigure}[b]{0.32\textwidth}
         \centering
         \includegraphics[width=\textwidth]{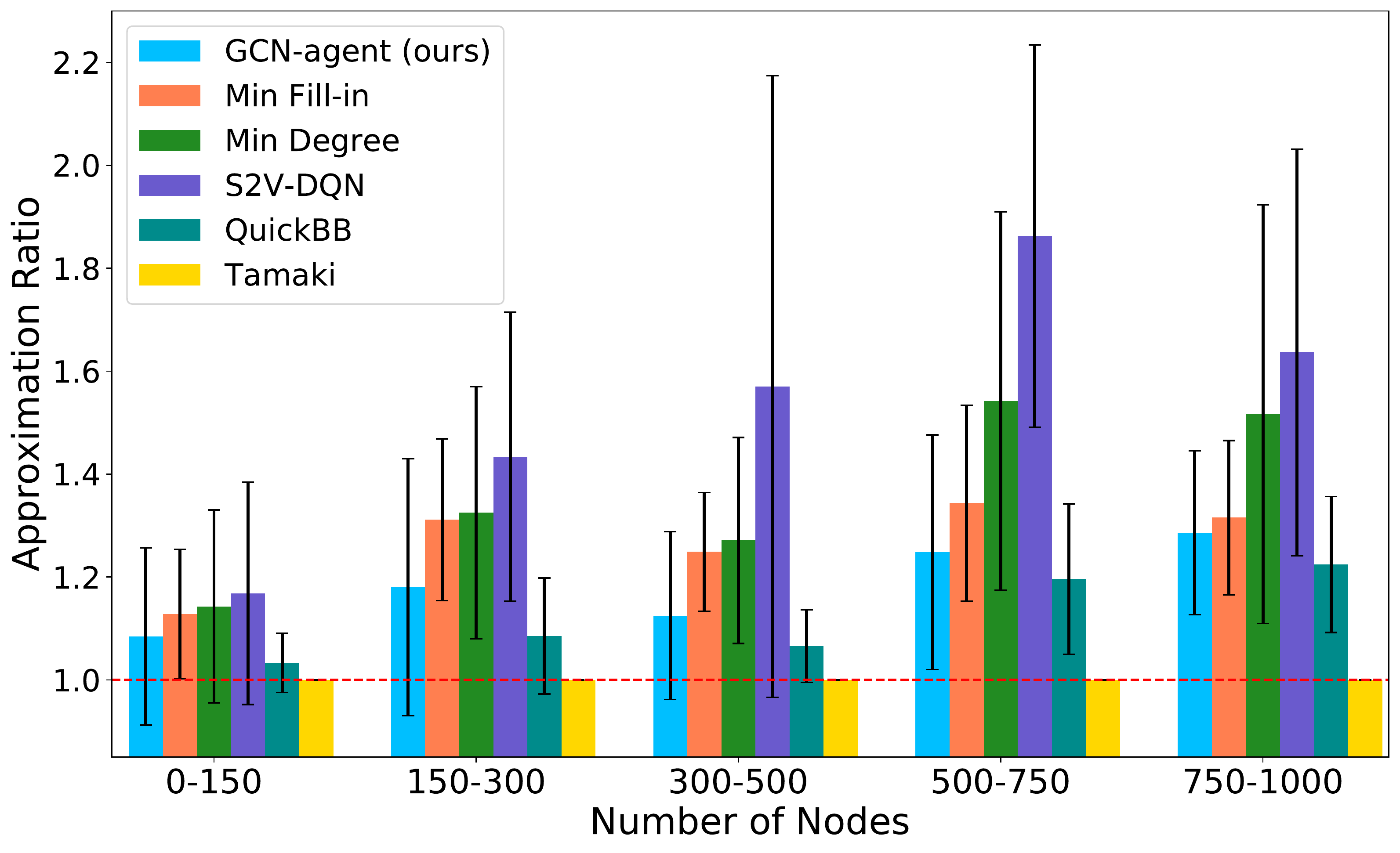}
         \caption{PACE2017}
         \label{fig:pace_bar}
     \end{subfigure}
     \hfill
     \begin{subfigure}[b]{0.32\textwidth}
         \centering
         \includegraphics[width=\textwidth]{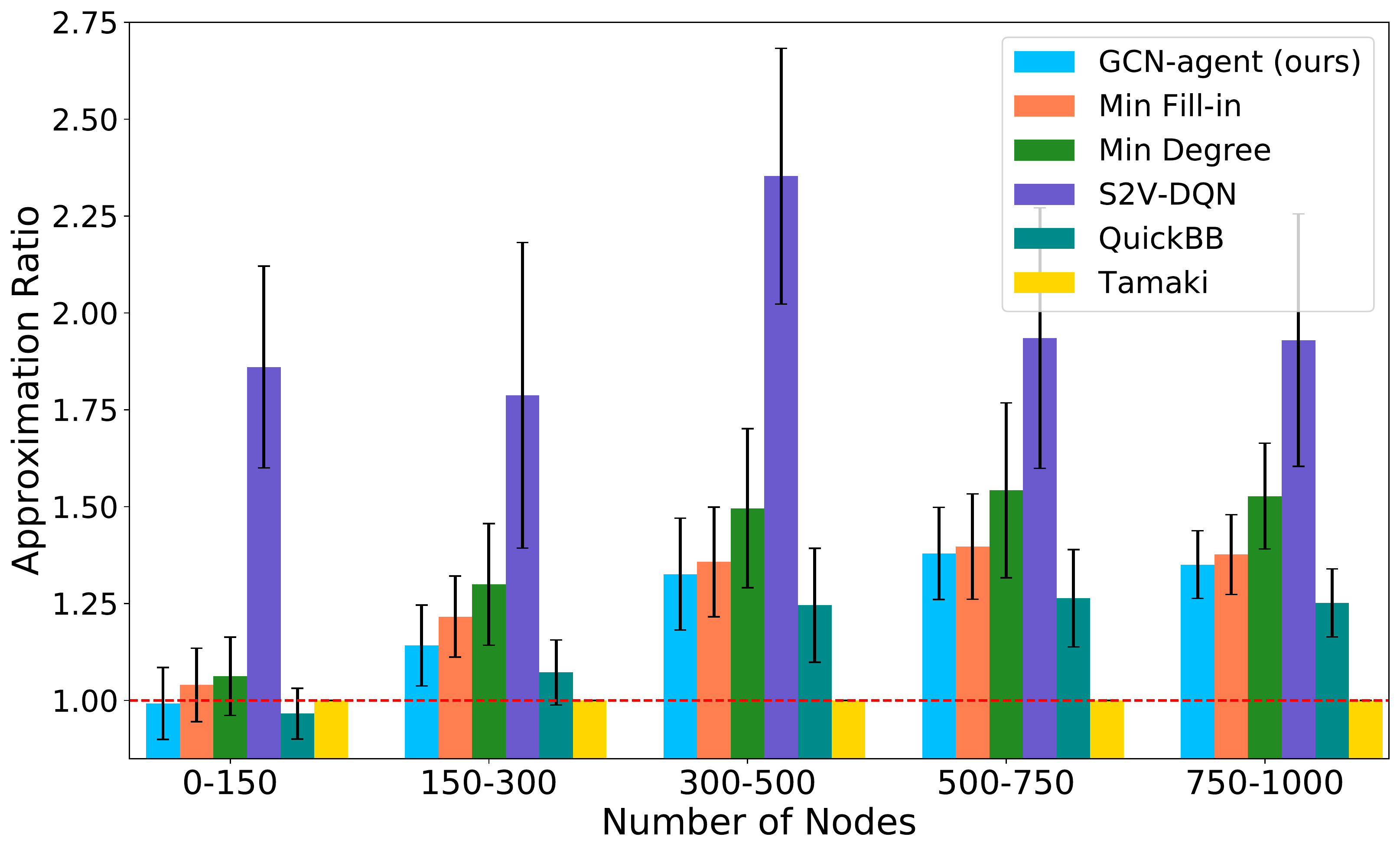}
         \caption{Circuits}
         \label{fig:circuit_bar}
     \end{subfigure}
    \hfill
        \caption{Approximation Ratio (less is better) of different methods averaged over all graphs in the respective dataset. The red line shows the AR compared to Tamaki solver with 30 minutes time bound. The result of the agent (trained on the single graph) slightly better than all greedy human-designed heuristics.}
        \label{fig:three_bars}
        \vspace{-13.8pt}
\end{figure*}
\subsection{Choice of the Training Graph}
In this series of experiments, we would like to check how the choice of training dataset affects the performance of the resulting agent. In the deep RL, it is common to train the agent on a huge number of problem instances and to expect an increase in the quality of the model from a larger number of training examples. 
We check this intuition in the following experiment. First, we train the agent on a single graph. We choose one ER graph with 68 nodes and one graph from the PACE dataset ("he064"). Next, we train on different ER and PACE graphs with up to 200 nodes. We randomly select a graph for every training step.
The performance of the agents is compared to the same validation set of Erd\H{o}s–R\'enyi graphs. The results are shown in Figure \ref{fig:selection_agent}. It turns out sufficient to train our agent for $\sim 10$ epochs (around 20 minutes). The S2V-DQN requires around $100$ epochs to find a good policy. Still, the quality of the DQN-based agent is lower. The comparative results are listed in Table~\ref{tab:s2v_over_our}. 

 Surprisingly, we find that the score of the agent does not significantly depend on the source of the training graph, despite the graphs, which we use, have quite different structures. Also, the size of the training set does not significantly affect final accuracy but rather resulted in longer training times.
 This fact may be attributed to the inefficiency of the agent or to the uniform structure of the TD problem already for average-sized graphs (with 50 nodes).
 We admit that this intriguing fact may need additional investigation. To simplify the reproducibility of our experiments, we choose Erd\H{o}s–R\'enyi graphs for further tests.

 In another set of experiments, we consider training on graphs of different sizes. It is known that if the action space is large, the RL methods may have problems with convergence due to the significant variance of the gradients. It is hence desirable to keep the training graph size small (for example, less than 100 nodes); however, training on larger instances may produce agents with better performance. To check the influence of the training graph size on the accuracy of the learned heuristic, we train separate agents on ER graphs with 10 to 210 nodes. The dependence of the AR is shown in Figure~\ref{fig:selection_graph_size}.

 As can be seen from Figure~\ref{fig:selection_graph_size}, the accuracy of the GCN-agent depends on the number of nodes only slightly. Also, it can be noted that better accuracy is obtained for the graphs with 70 and 130 nodes, which is approximately the GCN feature size or twice feature size. For further experiments, we select a graph with 70 nodes.
 
\subsection{Comparison with Other Solvers}
In this section, we compare the performance of our agent to other methods on the graphs of the different sizes and structures. The agent is trained on the ER graph with 70 nodes. The results of all experiments are summarized in Figure~\ref{fig:three_bars}.
The learned solver usually does not reach the accuracy of specialized exact algorithms but outperforms greedy heuristics. It is interesting to see that one can learn a relatively useful heuristic using only a single graph.


\begin{table}[h!]
\centering
\begin{tabular}{llll}
\toprule
 Method & \begin{tabular}[c]{@{}l@{}} Approx. \\ Ratio\end{tabular}  $\big\downarrow$&  \begin{tabular}[c]{@{}l@{}} Ratio  \\ Max. \end{tabular} & \begin{tabular}[c]{@{}l@{}}Avg. \\ Time, \\ sec \end{tabular}   \\
\midrule
\citet{tamaki2019positive} & 1.0 & 1.0 & 17.01  \\
QuickBB & 1.09 $\pm$ 0.11 & 1.41 & 1617.27  \\
Min-Fill & 1.24 $\pm$ 0.22 & 1.78 & 153.52   \\
Min-Degree & 1.31 $\pm$ 0.23 & 2.13 & 0.04  \\
GCN-Agent (ours) & 1.16 $\pm$ 0.16 & 1.44 & 10.93  \\
S2V-DQN & 1.45 $\pm$ 0.44 & 1.79 & 0.9  \\
Random agent & 1.84  & -& -  \\
\bottomrule \\
\end{tabular}
\caption{A summary of the agent's performance on the PACE2017 dataset. The values are averaged over all graphs with sizes from 10 to 1000 nodes. The GCN solver performs better than all greedy heuristics and has lower time-to-solution. }
\label{table:erdoc_pace_scores}
\end{table}
 
 The quality of the agent's solution deteriorates as the size of the test graphs grows, but usually slower than the quality of the solution found by greedy heuristics.
 
 Our Actor-Critic based agent performs well of all three datasets, despite being trained on the single ER graph, and achieves the significant level of generalization. However, a DQN agent trained on the ER graph generally performs poorly on graphs with different structures.
 This observation may be explained by the fact that DQN-based methods select the policy greedily, which may lead to poor performance if the distributions of the training and test data are different. However, once a high-quality deterministic policy is found, the DQN-based method is more efficient than the Actor-Critic, which may produce the policy with some portion of stochastic behavior. Once again, our agent outperforms all greedy heuristics on the graphs with different structures and the number of vertices.

The results on the PACE2017 dataset, which we consider the most representative, are summarized in Table~\ref{table:erdoc_pace_scores}. We note that in addition to the high accuracy comparing to greedy heuristics (both on average and in the worst case), the RL-based heuristic has very competitive time-to-solution, which is essential for solving NP problems. Furthermore, this time can be significantly decreased, as sampling can be trivially performed in parallel.

At last, we would like to discuss the ways to improve our approach, as the agent still can rarely reach the accuracy of specialized algorithms. Recall that we formulated TD as a search for the optimal order of nodes. At each step, the reward is closely related to the degree of the vertex. Hence the policy/value networks are guided to use the local structure of the graph. In contrast, the current state of the art algorithms for TD are based on finding vertex separators, which are global structures relating distant parts of the graph. We would speculate that the TD problem is hard in the ordering formulation. Despite the intuitive definition of the reward, it provides poor feedback to the RL algorithm during the search of the solution. Reformulation of the RL task may be a way to produce better heuristics.

Nevertheless, it is fascinating that a high-quality heuristic can be produced using a single graph, and that this heuristic agent can produce a near-optimal solution of the combinatorial problem in a fixed time.

\begin{figure}[t!]
    \centering
    \vspace{-10pt}
    \includegraphics[width=0.48\textwidth,height=6.5cm]{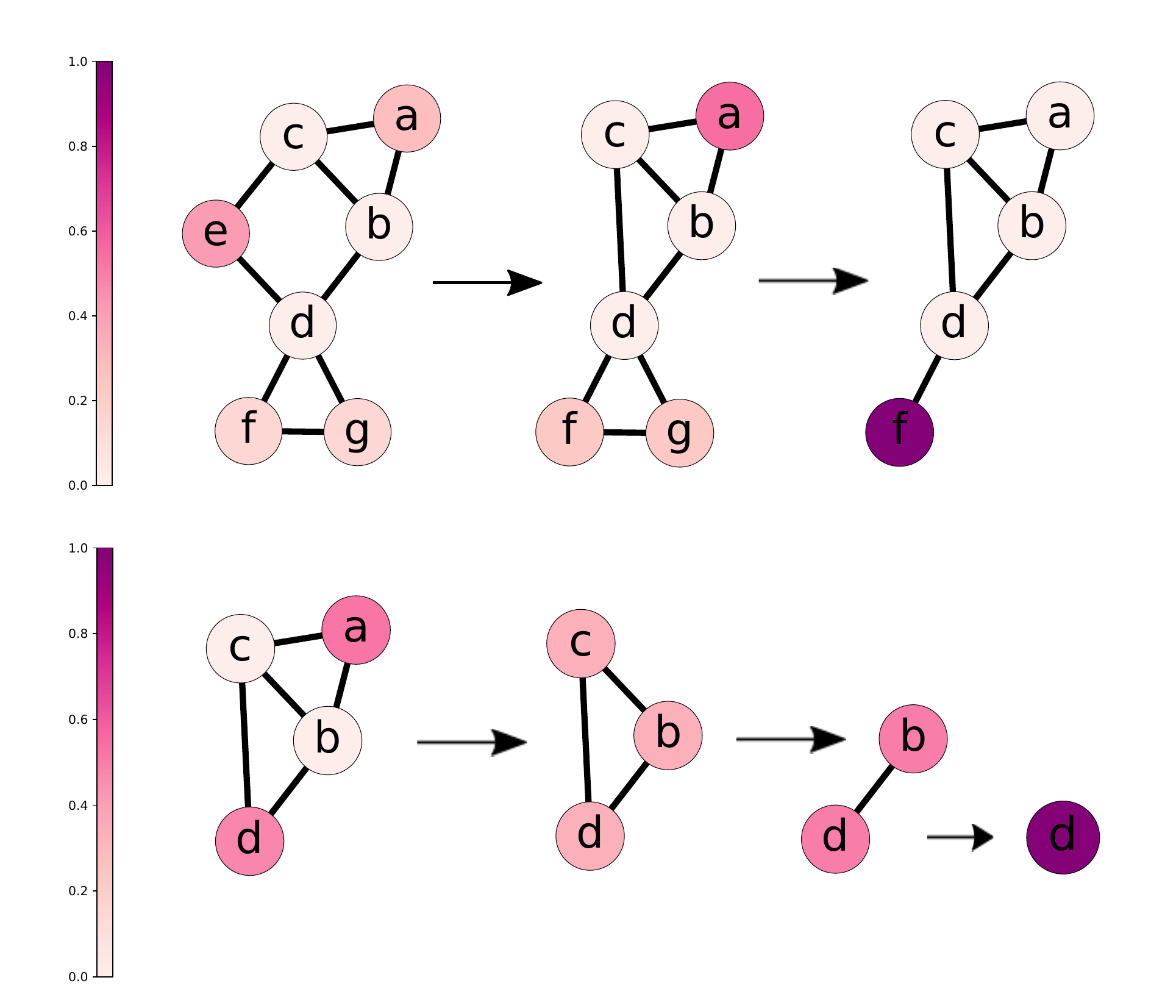}
    \caption{Elimination procedure of an agent. The probabilities of actions are shown in color. At each step, the next node is sampled according to the probability. Starting from the fifth step, when a final clique is encountered, the distribution is uniform.}
    \label{fig:agent_decision}
    \vspace{-13.8pt}
\end{figure}

\subsection{Agent Decision Making}
In this section, we analyze the agent's decision-making process in order to get insights into the structure of the TD problem.

As mentioned earlier in all experiments, we use sampling to get the candidate solutions and select the one with the best score. We use a small sample of 10 trajectories, which is a stronger result comparing to other works on neural-based heuristics, where samples of size 1000 are common \citep{Weilling_routing18}. 
The efficiency with a small sample size suggests that the agent learns a distribution that may be close to the "true" distribution of solutions. 

Another evidence in support of the high quality of the learned distribution is the behavior of the agent on complete graphs. A fully connected graph with $n$ nodes has treewidth $n - 1$, and all elimination orders on it are equivalent. The policy learned by our agent produces a uniform distribution on the complete graphs (after they appeared during the elimination process), as shown in Figure~\ref{fig:agent_decision}. 
Also, we provide additional numerical experiments in Appendix \ref{appendix:secon},  which clarify the distribution over remaining cliques in the agent decisions.

%% file: parts/related_works.tex
\section{Related Works}
In recent years, several pioneering works concentrated on applying neural networks to learn heuristics for NP-hard problems due to the desire to get an approximate near-optimal algorithm without using domain-specific knowledge. 
Pointer Network \citep{Vinyals16} is the first neural-based sequence-to-sequence model applied in this context.
The authors used attention over the input architecture to solve permutation-based combinatorial optimization problems, such as Euclidean TSP.
Another work by \citet{BelloPLNB16} improved the heuristics for TSP using the Actor-Critic \citep{Konda:2002} algorithm. 
A more recent article by \citep{NIPS2018_8190} focuses on the Vehicle Routing Problem, which is also defined on the 2D Euclidean space, similar to TSP.
Another approach for 2D TSP variants was proposed in \citep{Weilling_routing18}.
The authors created a very efficient way to solve small TSP instances using modern techniques from machine translation. Still, their method is not very effective numerically as more than 1000 agent trajectories were sampled to find the solution. 

The main drawback of previous approaches in the context of TSP is an explicit utilization of the Euclidean structure of the problem, which prevents the application of these techniques to other combinatorial problems on graphs. Recently  \citep{StructureVec} proposed a graph embedding model that can be naturally applied to arbitrary graphs. In a subsequent paper \citep{DaiKZDS17}, the authors used these embeddings to create an RL-based solution for Minimum Vertex Cover, Euclidean TSP, and Maximum Cut problems. They demonstrated that the agent could generate near-optimal solutions. Unfortunately, their graph embeddings should be trained on a large number of graphs, which can be computationally expensive. The method the authors used to train the agent is based on the DQN \citep{dqn} and may produce a greedy result, which can be a drawback in some applications.


One of the best results in finding neural-based heuristics was presented by \citep{Kotlun18}. Their approach is based on the combination of neural graph representation and a classical heuristic. The authors use GCN with a guided tree-search in a supervised setting. The only drawback of their method is the need for supervised learning for the embedding of the input graph, i.e., the method requires a large number of solved NP problem instances for training. Summarizing previous works, it seems that no universal purely RL-based technique for NP-problems exists to date.

%% file: parts/conclusion.tex
\section{Conclusion}
This paper presents the Tree Decomposition problem as a new task for representation learning. We propose a model, which can directly learn how to solve this combinatorial optimization problem using a \emph{single graph} for training. The training procedure also can be performed using a large set of graphs, but this work aims to show the generalization ability of a simple agent trained on a single graph. We show that a learnable heuristic can beat greedy, manually designed ones. Our method can generalize to large problem instances without significantly sacrificing the quality of the solution and without a large increase in computational time. 
We extensively verify the performance of the learnable heuristic using three datasets containing graphs with very different structures.

Our preliminary results suggest that this approach is a good starting point for learning heuristics on the graph-structured data. 
Outperforming specialized algorithms is a challenge in our setting. We are planning to extend our results to different NP-problems, which can be formulated in the sequential decision framework.
The perspective direction of research is a combination of RL-based heuristics with local-search methods. 

%% file: parts/appendix.tex
\begin{appendices}
\section{Reconstruction of the Tree Decomposition from an Ordering of Vertices\label{app:td_from_order}}
Here we provide a procedure to produce a tree decomposition of a graph $G = (U, E)$ given an ordering of vertices $\pi: U \longrightarrow 1, \ldots, |U|$. An extensive presentation of the procedure with proofs can be found in \citep{clique_trees} and in \citep{bodlaender1994tourist}. Here we merely list our variant of the algorithm for the reader's reference.

We denote the neighborhood of node $u$ by $\mathcal{N}(u)$. We present here an algorithm that builds a tree decomposition in the breadth-first search fashion starting from the leaves of the tree. Given an ordering of vertices $\pi$ of a graph $G = (U, E)$, one can build its tree decomposition $F = (B, T)$ with the following procedure:
\begin{algorithm}[t]
  \caption{Building tree decomposition from the elimination order}
  \label{code:build_td}
  \begin{algorithmic}[1]
    \Require $G = (U, E), \pi: U \rightarrow N, ~~ \pi = \{(u_{i}, i)\}_{i=1}^{|U|}$
    \Ensure $F = (B, T)$
    \Statex
    \Function{Build\_tree\_decomposition}{$G, \pi$}
    \State leaf\_bags $\gets \emptyset$
    \For{$i \in [1, \ldots, |U| - 1]$}
        \State $u \gets \pi^{-1}(i)$
        \For{$w, x \in \mathcal{N}(u)$}
            \State $E \gets E \cup (w, x)$
        \EndFor
        \If{$\mathcal{N}(u) \neq \emptyset$}
            \State $b = \mathcal{N}(u) \cup u$
        \EndIf
        \State $U \gets U \setminus u$
        \For{$l ~ \text{in leaf\_bags} $}
            \If{$b\subset l$}
              \State $b \gets l$
              \State break
            \EndIf
        \EndFor
        \For{$l ~ \text{in leaf\_bags} $}
            \If{$u \in l \cap b$ \textbf{and} $b \not\subset l$}
                 \State leaf\_bags $\gets$ leaf\_bags $\setminus ~ l$
                \State $B \gets B \cup b$
                \State $T \gets T \cup (l, b)$
            \EndIf
        \EndFor
     \EndFor
  \EndFunction
  \end{algorithmic}
\end{algorithm}
The algorithm runs an elimination procedure and builds a TD simultaneously. The neighborhoods of each node in the elimination order $\pi$ are bags in the TD. Here we skip bags, which are subsets of their child bags. 

The idea behind the algorithm is that parent bags in the tree emerge later in the elimination sequence than their children. If there is an intersection between two bags, then an edge has to exist in the tree. The algorithm keeps a queue of current leaf bags and checks the next bag against this list. If the parent of a leaf bag is found, the child is removed from the queue, and an edge is introduced in the tree graph $F$. 

\section{Entropy in agent decision \label{appendix:secon}}
To get a better insight into the structure of the learned policy, we study the behavior of the entropy. At each step of the elimination trajectory, $\pi$ the entropy of policy $\Pi (u|s)$ is defined as:
\begin{equation*}
    H(s) = -\sum_u \Pi (u|s) \log \Pi (u|s)
\end{equation*} 
As the size of the action space is not the same for different states in our problem, we define a  normalized entropy:
\begin{equation}
\hat{\mathcal{H}}(s) = \frac{H(s)}{\log |U_s|}     
\end{equation}
Here $|U_s|$ is the size of the action space for state $s$. 

\begin{figure*}[ht!]
    \centering
    \begin{subfigure}[b]{.25\linewidth}
    \includegraphics[width=\linewidth]{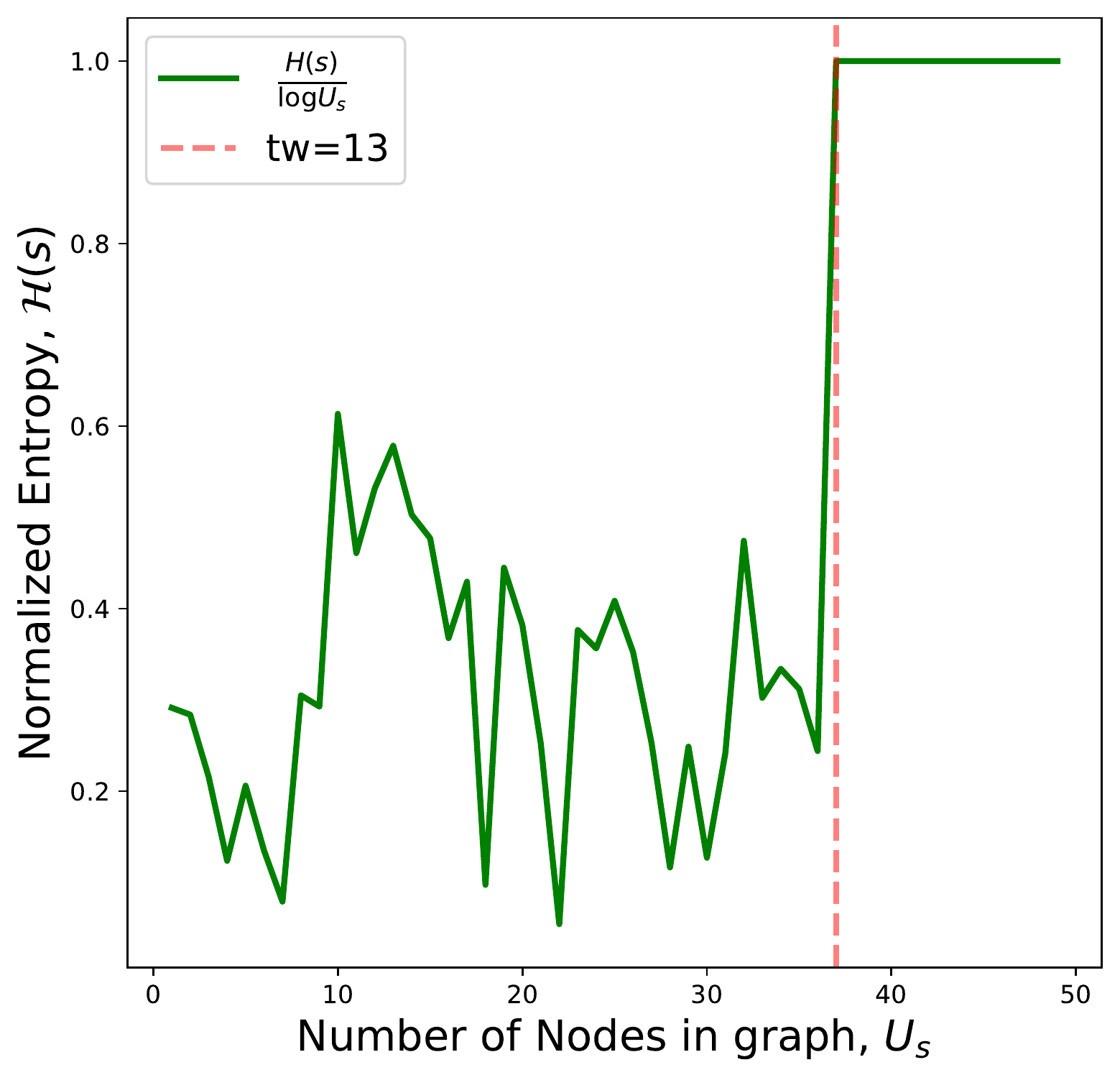}
    \caption{\label{fig:entropy_erdos0}ER, $|U|=50$,\\ $AR=1.0$.}
\end{subfigure}
\qquad
\begin{subfigure}[b]{.25\linewidth}
    \includegraphics[width=\linewidth]{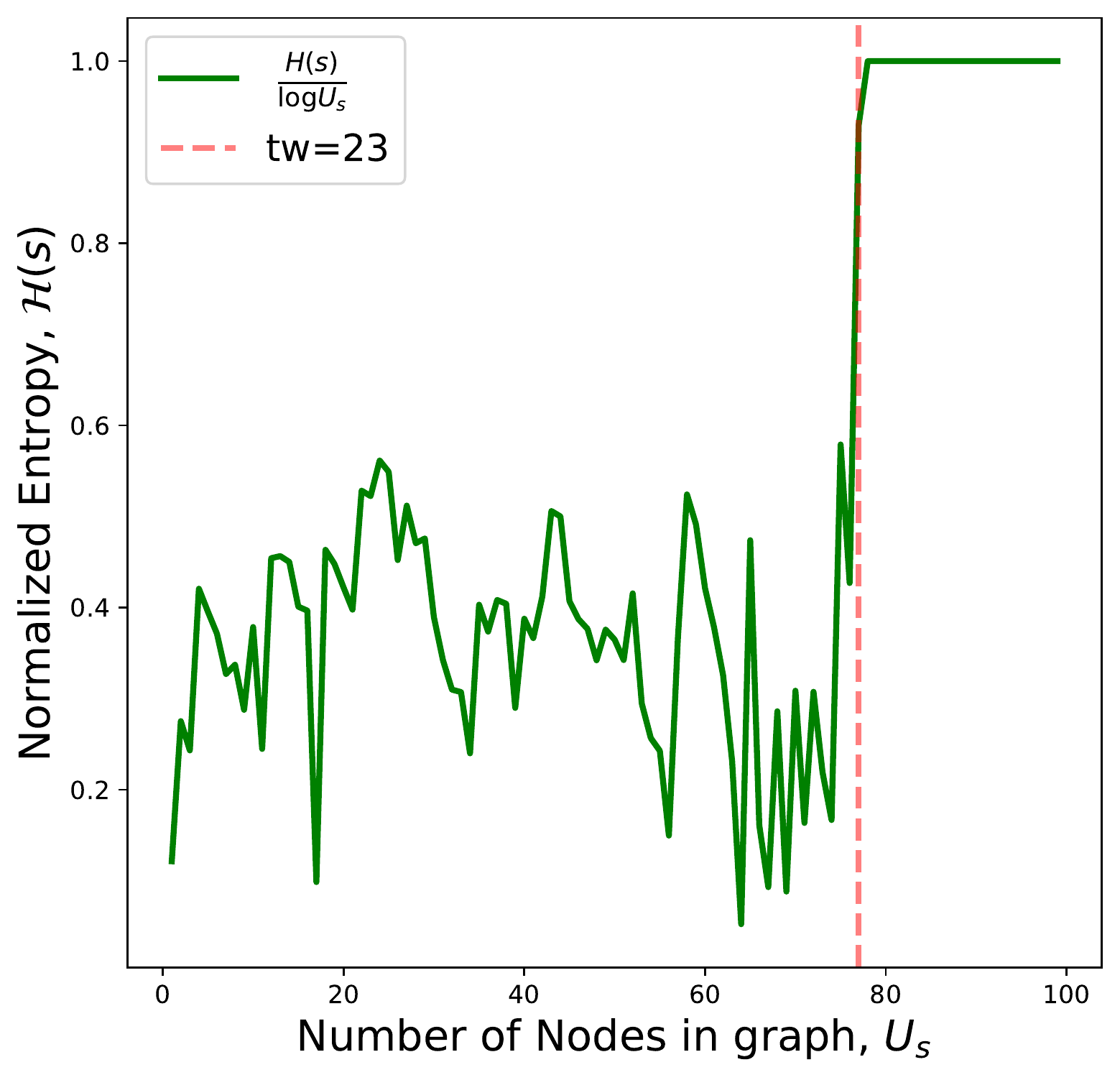}
    \caption{\label{fig:entropy_erdos1}ER, $|U|=100$,\\ $AR=1.06$.}
\end{subfigure}
        \vskip\baselineskip
    \begin{subfigure}[b]{.25\linewidth}
    \includegraphics[width=\linewidth]{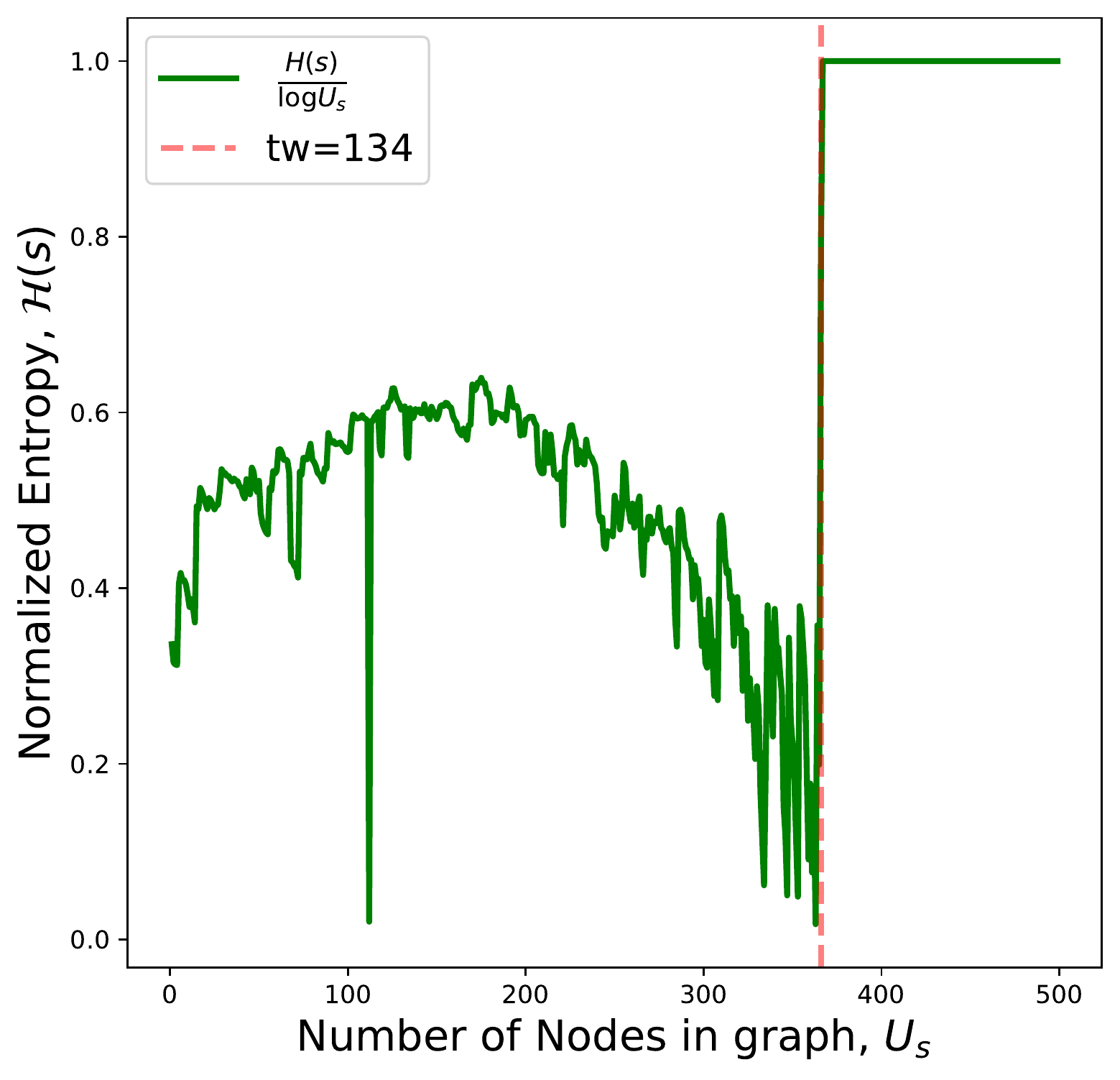}
    \caption{\label{fig:entropy_erdos2}ER, $|U|=500$,\\ $AR=1.0$.}
\end{subfigure}
\qquad
\begin{subfigure}[b]{.25\linewidth}
    \includegraphics[width=\linewidth]{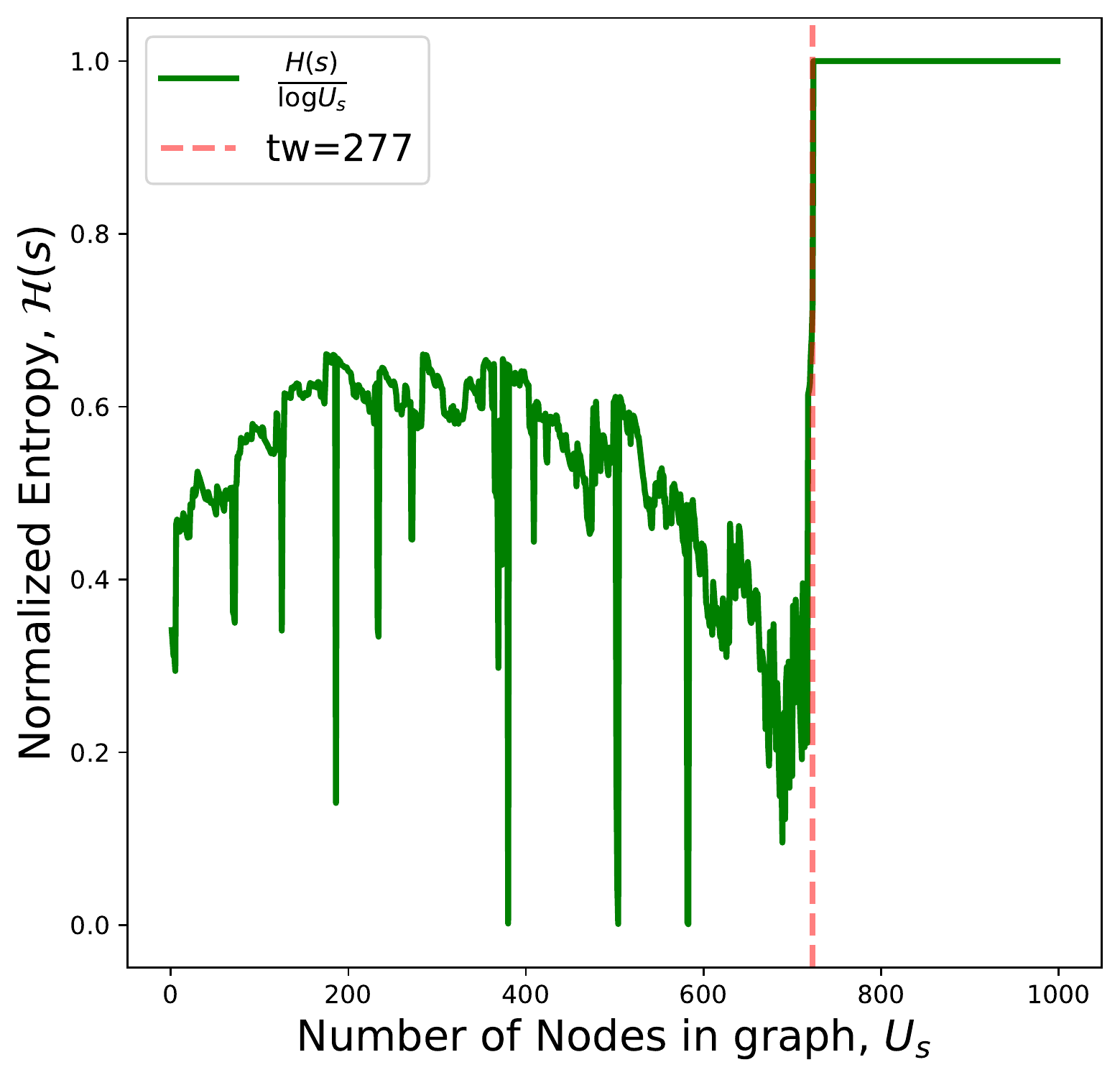}
    \caption{%
\label{fig:entropy_erdos3} ER, $|U|=1000$,\\ $AR=1.08$.}
\end{subfigure}
        \vskip\baselineskip
    \begin{subfigure}[b]{.25\linewidth}
    \includegraphics[width=\linewidth]{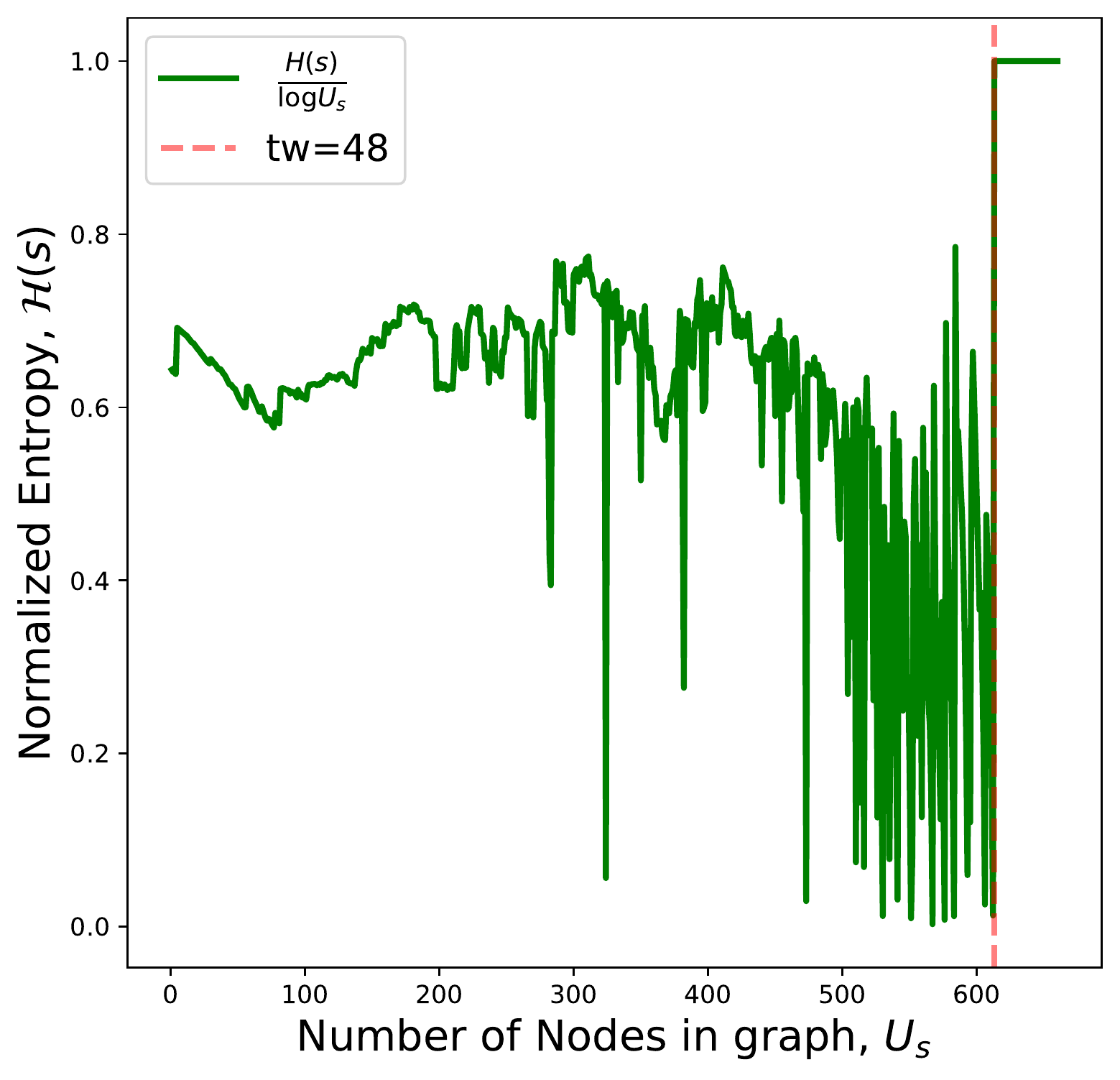}
    \caption{\label{fig:entropy_pace0}PACE "he030.gr",\\ $|U|=661$, $AR=1.0$.}
\end{subfigure}
\qquad
\begin{subfigure}[b]{.25\linewidth}
    \includegraphics[width=\linewidth]{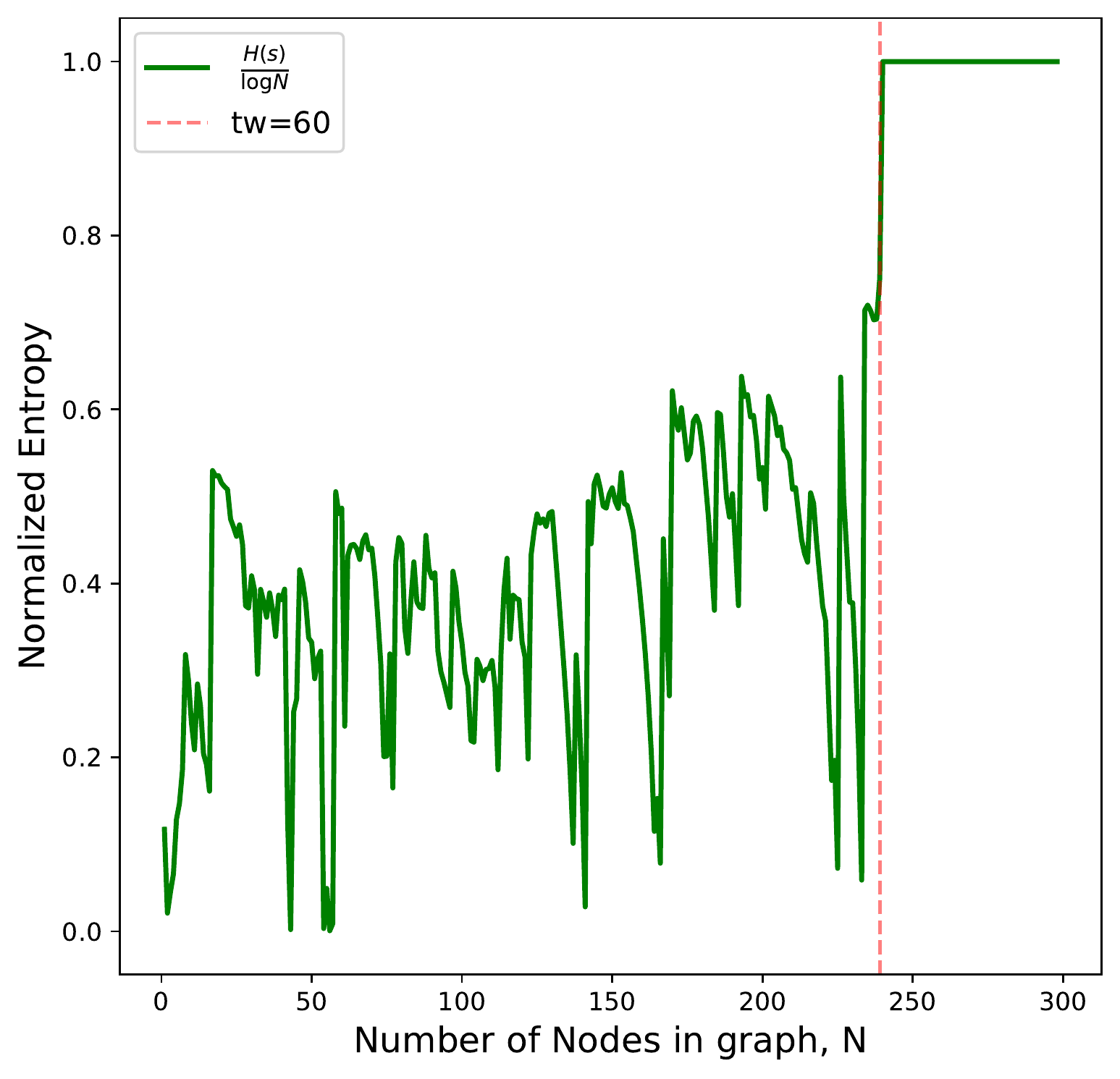}
    \caption{%
\label{fig:entropy_pace1}%
 PACE "he043.gr",\\ $|U|=299$, $AR=1.1$.}
\end{subfigure}
        \vskip\baselineskip
\begin{subfigure}[b]{.25\linewidth}
    \includegraphics[width=\linewidth]{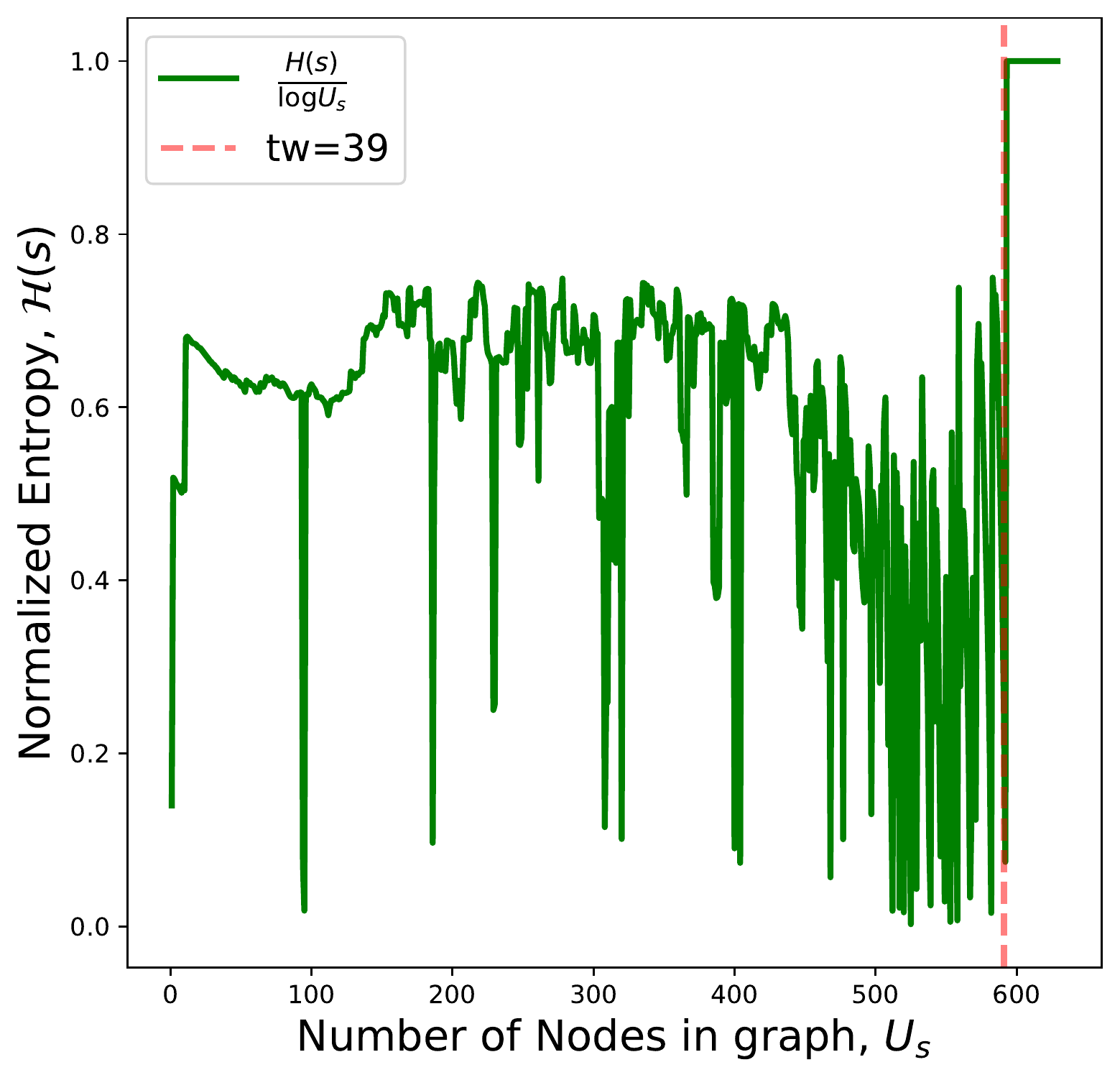}
    \caption{%
\label{fig:entropy_pace2}%
PACE "he031.gr",\\ $|U|=630$, $AR=1.03$.}
\end{subfigure}
\qquad
\begin{subfigure}[b]{.25\linewidth}
    \includegraphics[width=\linewidth]{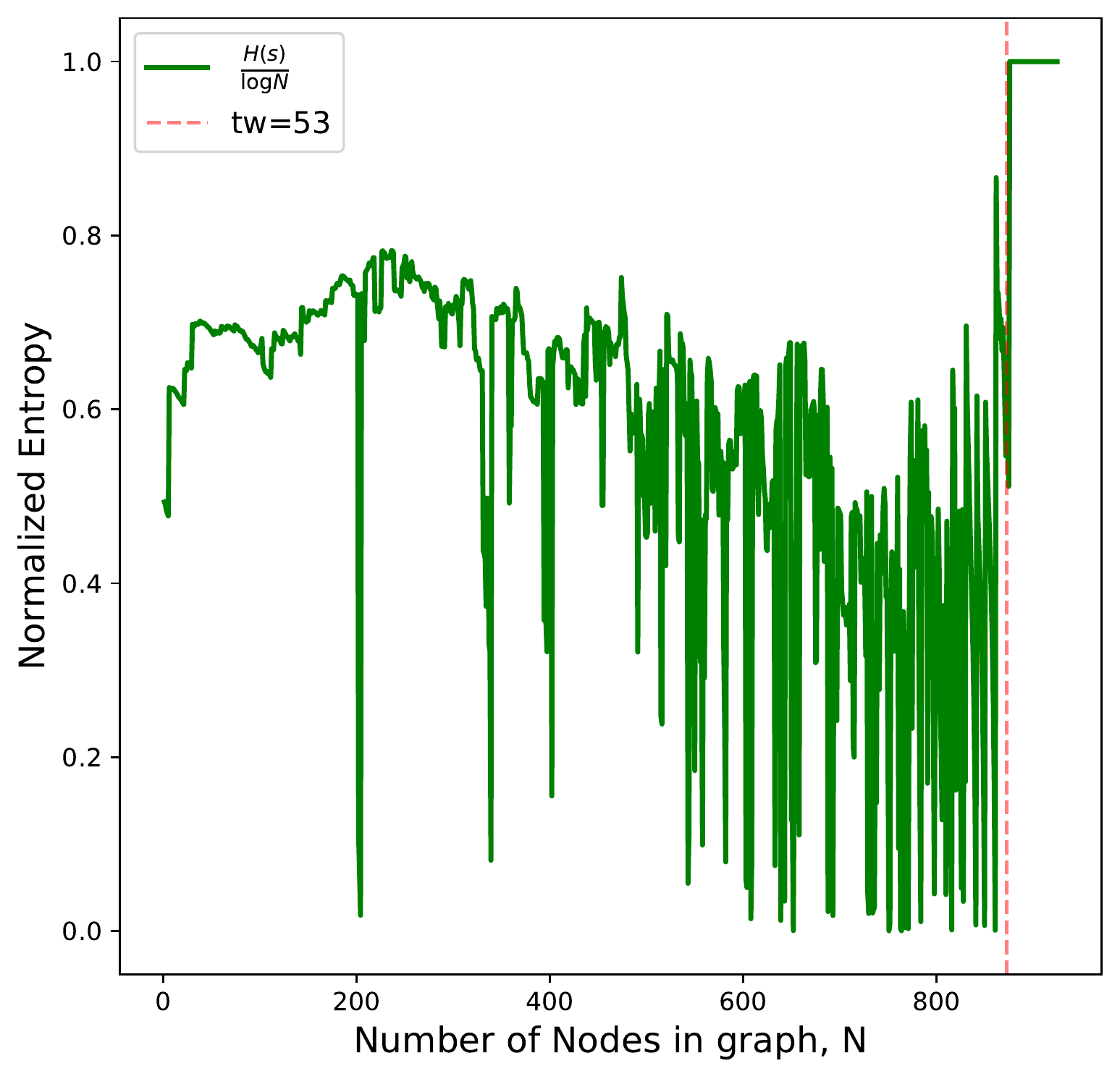}
    \caption{%
\label{fig:entropy_pace3}PACE "he019.gr",\\ $|U|=926$, $AR=1.12$.}
\end{subfigure}
\caption{Normalized entropy $\mathcal{\hat{H}}(s)$ evaluated on every step of elimination procedure for four Erd\H{o}s–R\'enyi's graphs and four randomly sampled PACE2017 graphs (names of the graphs are provided in quotes).
Every caption contains information about the AR for the first ordering sampled from the policy. $|U|$ is the number of nodes in the corresponding graph. The vertical red line shows the treewidth (the line intersects $x$ axis at $|U| - tw$).}
\label{fig:entropy_result}
\end{figure*}

Note that if the distribution is uniform, e.g. if $\Pi(u|s) = \frac{1}{|U_{s}|}$, then $H(s) = \log|U_s|$ and $\hat{H}(s) = 1$. 
We plot normalized entropies for several trajectories on different graphs in Figure \ref{fig:entropy_result}.

A clear pattern is seen in the learned policies. The agent eliminates vertices in such a way that the largest clique appears at the end of the ordering.
If the treewidth of the order found by the agent is $k$, then the size of the largest clique in the elimination sequence is also $k$.
After this last clique is formed, the order of elimination is irrelevant (it does not increase the treewidth). This fact is reflected by the plateaus on the right side of the normalized entropy graphs (the normalized entropy of the uniform distribution is 1).
Also, we can see in most cases the normalized entropy is far from zero. Consequently, there are multiple choices that lead to the orderings of the same treewidth according to the policy. We assume that the learned policy is close to the distribution of the solution, which is a plausible assumption considering low values of AR and a correct behavior of the agent on fully connected graphs. 
\onecolumn

\end{appendices}